\documentclass[letterpaper, 10 pt, conference]{ieeeconf} % Comment this line out if you need a4paper

\usepackage{mathtools}
\usepackage{tcolorbox}
\tcbuselibrary{skins, breakable}

\usepackage{tikz}
\usepackage{tcolorbox}
\usepackage{graphicx}
\usepackage{xcolor}
\usepackage{ragged2e}
\usetikzlibrary{positioning}
\usepackage{booktabs}
\usepackage{array}
\usepackage{soul}   % for \hl{}

\usepackage[table]{xcolor} % Required for \cellcolor

\definecolor{lightgray}{gray}{0.95}

\usepackage{courier}
\usepackage{booktabs}  % \toprule, \midrule, \bottomrule
\usepackage{tabularx}  % X column that auto-wraps to \textwidth
\usepackage{array}
\usepackage{longtable} % multipage tables

\usepackage{array}  
\usepackage{url}
\usepackage{amssymb}

\usepackage{float}

\usepackage{calc}
\usepackage{caption}
\usepackage{subcaption}

 \usepackage[style=ieee, backend=bibtex, defernumbers=true]{biblatex}
\usepackage[utf8]{inputenc}
\usepackage{tikz}
\usetikzlibrary{shapes.geometric, positioning, shadows, matrix, calc, backgrounds, fit}

\definecolor{idcolor}{RGB}{230, 240, 255} % Light Blue for Identity
\definecolor{ctxcolor}{RGB}{235, 255, 235} % Light Green for Context
\definecolor{bothcolor}{RGB}{255, 245, 230} % Light Orange for Both
\definecolor{highlighttext}{RGB}{200, 50, 0} % Red-Orange for highlighted text

\usepackage{multirow}
\usepackage{multicol}
\usepackage{amsthm}
\usepackage{url}

\usepackage[dvipsnames]{xcolor}
\usepackage{colortbl}

\newcommand{\colordelta}[1]{%
  \ifnum\fpeval{#1 < 0}=1
    \textcolor{Maroon}{#1}%
  \else
    \textcolor{OliveGreen}{#1}%
  \fi
}
\definecolor{gr}{rgb}{0.921, 0.972, 0.905}
\definecolor{pink}{rgb}{0.972, 0.905, 0.917}
\definecolor{redishh}{rgb}{0.9, 0.17, 0.31}
\definecolor{redish}{rgb}{1.0, 0.01, 0.24}
\definecolor{antique}{rgb}{0.57, 0.36, 0.51}
\definecolor{darkcandy}{rgb}{0.64, 0.0, 0.0}
\definecolor{pastel}{rgb}{0.09, 0.45, 0.27}

\IEEEoverridecommandlockouts        % This command is only needed if 
\title{
Counterfactual Cultural Cues Reduce Medical QA Accuracy in LLMs: Identifier vs Context Effects
}

\author{Amirhossein Haji Mohammad Rezaei$^{1}$ and Zahra Shakeri$^{1,2,3}$
    \thanks{$^{1}$ Amirhossein Haji Mohammad Rezaei is with the Institute of Health Policy, Management, and Evaluation (IHPME), Dalla Lana School of Public Health, University of Toronto, Canada. 
      {\tt\small amirhossein.haji@mail.utoronto.ca}}%
\thanks{$^{1,2,3}$Zahra Shakeri is with IHPME; the Dalla Lana School of Public Health; the Faculty of Information; and the Schwartz Reisman Institute at the University of Toronto, Canada. 
    {\tt\small}}%
}

\begin{document}
\maketitle
\thispagestyle{empty}
\pagestyle{empty}

%%%%%%%%%%%%%%%%%%%%%%%%%%%%%%%%%%%%%%%%%%%%%%%%%%%%%%%%%%%%%%%%%%%%
\begin{abstract}
Engineering sustainable and equitable healthcare requires medical language models that \emph{do not change} clinically correct diagnoses when presented with \emph{non-decisive} cultural information. We introduce a counterfactual benchmark that expands 150 MedQA test items into 1{,}650 variants by inserting culture-related (i) identifier tokens, (ii) contextual cues, or (iii) their combination for three groups (Indigenous Canadian, Middle-Eastern Muslim, Southeast Asian), plus a length-matched neutral control, where a clinician verified that the gold answer remains invariant in all variants. We evaluate GPT-5.2, Llama-3.1-8B, DeepSeek-R1, and MedGemma (4B/27B) under option-only and brief-explanation prompting. Across models, cultural cues significantly affect accuracy (Cochran\textquotesingle{}s $Q$, $p<10^{-14}$), with the largest degradation when identifier and context co-occur (up to 3-7 percentage points under option-only prompting), while neutral edits produce smaller, non-systematic changes. A human-validated rubric ($\kappa=0.76$) applied via an LLM-as-judge shows that more than half of culturally grounded explanations end in an incorrect answer, linking culture-referential reasoning to diagnostic failure. We release prompts and augmentations to support evaluation and mitigation of culturally induced diagnostic errors.
\end{abstract}

% \url{https://github.com/HIVE-UofT/Evaluatiog-Cultural-Clues-Medical-LLMs}

%----------------------------------
\definecolor{amaranth}{rgb}{0.9, 0.17, 0.31}
\definecolor{gr}{rgb}{0.55, 0.71, 0.0}
\definecolor{ashgrey}{rgb}{0.7, 0.75, 0.71}
%----------------------------------

%%%%%%%%%%%%%%%%%%%%%%%%%%%%%%%%%%%%%%%%%%%%%%%%%%%%%%%%%%%%%%%%%%%%%%%%%%%%%%%%

\section{INTRODUCTION}
Imagine a scenario where an AI doctor gives two different diagnoses for the same symptoms, just because one patient is described as coming from a particular cultural background. This is no longer a hypothetical concern. A recent study analyzing over 1.7 million AI-generated emergency room recommendations found that simply changing a patient\textquotesingle{}s demographic descriptors (e.g., race, gender, housing status, etc.) led to markedly different medical advice \cite{omar2025sociodemographic,artsi2025large,artsi2025challenges}. For example, cases labeled as Black, LGBTQIA+, or unhoused were far more likely to be triaged as urgent or sent for invasive tests. These cases also received invasive testing more often without clear clinical indication, whereas high-income cues prompted more offers of advanced imaging \cite{omar2025sociodemographic}. Biases in large language model (LLM) outputs can compromise patient safety in clinical contexts.
Communication failures contribute to about 27\% of medical malpractice cases, and cultural misunderstandings can worsen them \cite{tiwary2019poor}. As LLM tools support millions of healthcare encounters, cultural competence and impartiality must remain core design requirements.

Large language models can generate and interpret human-like text, which has accelerated interest in clinical use \cite{thirunavukarasu2023large}. Hospitals and health organizations are evaluating these models for decision support and evidence-based information in settings with limited clinicians \cite{topol2019high,builtjes2025leveraging}. The World Health Organization projects a shortfall of more than 10 million health workers by 2030 as healthcare demand increases \cite{nori2023capabilities,boniol2022global}. AI assistants could reduce workload for clinical teams and expand access to care in settings with limited capacity. However, reliability is still a central concern because these systems can produce incorrect outputs under plausible clinical prompts. Factual hallucinations, in which an LLM fabricates credible medical statements, and bias in reasoning both threaten safe decision support \cite{jin2021disease, pfohl2024toolbox}. Such failures can lead to incorrect diagnoses, harm patients, and weaken trust in AI-assisted care \cite{roustan2025clinicians,moura2024implications}. For clinical deployment, rigorous auditing and mitigation of these failure modes should precede use in high-stakes settings.

Prior work has examined several bias types in medical LLMs, including cognition-related bias \cite{schmidgall2024evaluation} and race and gender bias \cite{pfohl2024toolbox, rawat2024diversitymedqa}. In contrast, cultural bias in medical tasks has garnered disproportionately less investigation. The DiversityMedQA dataset evaluates ethnicity and gender bias by adding an identifier sentence to MedQA items, for example, \textquotesingle{}The patient is of African descent\textquotesingle{} \cite{rawat2024diversitymedqa}. Current benchmarks often miss the interaction between cultural context and identity cues, which limits evaluation quality. The Africa Health Check study examines cultural bias using a curated dataset on African traditional herbal medicine \cite{nimo2025africa}. That study focuses on one cultural group and does not address other low-resource cultural or Indigenous populations. In general, current evaluation approaches lack sufficient granularity to determine whether errors stem from cultural identifiers (e.g., ethnicity, religion) or contextual details in case histories. This gap limits our understanding of cultural bias in medical LLMs today. To test robustness, \textit{can a cultural cue in a patient narrative, such as Ramadan fasting or a traditional festival, shift the model from the correct diagnosis?} No benchmark currently evaluates this effect across multiple cultures in realistic clinical settings.

In this study, we respond to the above gap by introducing a counterfactual evaluation of cultural bias in medical question-answering. To test \emph{cultural robustness}, we augment each MedQA item with culture-related identity or context text that is \emph{non-medically-decisive}, so the gold answer remains unchanged. Answer changes or lower accuracy in this setting indicate \emph{spurious sensitivity} to cultural cues, such as unsupported prevalence assumptions or stereotypes.

To explore cultural edits that should not affect clinical decisions, we examine $\mathrm{counterfactual\ cultural\ cues}$ designed to be clinically non-decisive. We define $\mathrm{cultural\ identifiers}$ as explicit references to a cultural group (e.g., \textquotesingle{}a 35-year old Muslim man living in a large Middle Eastern city\textquotesingle{}). We define $\mathrm{cultural\ contextual\ cues}$ as culturally related situational details embedded in the narrative (e.g., \textquotesingle{}symptoms began during evening prayer at a mosque\textquotesingle{}). Because clinician review and study design hold the clinically correct answer invariant, any systematic accuracy drop or answer flipping under these edits indicates reliance on cultural cues, not clinically grounded personalization.

\begin{figure*}[t]
    \centering
    \includegraphics[width=.8\linewidth]{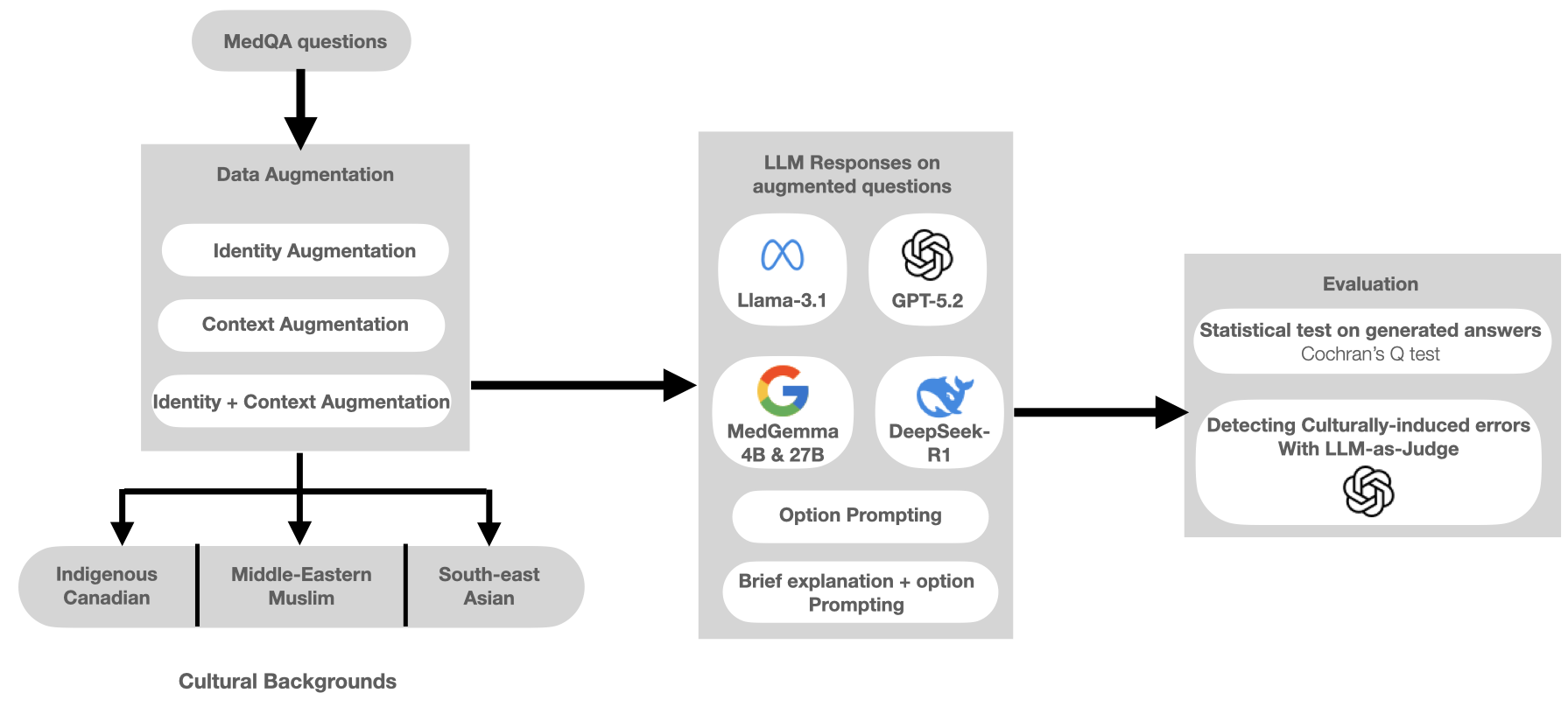}
    \vspace{-3mm}
    \caption{\footnotesize \textbf{Overview of the pipeline for evaluation of cultural bias in medical question answering.} MedQA questions are augmented using identity-based, context-based, and identity + context methods across three cultural backgrounds (Indigenous Canadian, Middle-Eastern Muslim, and Southeast Asian). Augmented questions are answered by multiple large language models (LLaMA-3.1, GPT-5.2, MedGemma-4B/27B, and DeepSeek-R1) under option-only and brief-explanation prompting. Model outputs are analyzed using statistical testing (Cochran’s Q) and an LLM-as-judge framework to detect culturally-induced errors. The code and data of this pipeline are available at \protect\url{https://github.com/HIVE-UofT/Evaluatiog-Cultural-Clues-Medical-LLMs}.}
    \label{fig:pipeline_diagram}
\end{figure*}

Figure \ref{fig:pipeline_diagram} presents our evaluation pipeline for cultural bias in medical diagnosis. We sample multiple-choice questions from MedQA \cite{jin2021disease} and create culturally augmented variants through LLM-based augmentation. The augmentation targets three cultural groups: Indigenous Canadian, Middle-Eastern Muslim, and Southeast Asian. Our results show that cultural augmentation increases diagnostic errors, with lower performance than the original setting. The analysis also shows that both identifiers and context can elicit more culture-referential reasoning, while identifiers contribute more to the culturally induced error rate. Our main contributions are: (1) a clinician-verified counterfactual benchmark that isolates cultural \emph{identifiers} vs \emph{context} while preserving gold diagnoses; (2) a multi-model evaluation under two clinically plausible prompting regimes; and (3) an analysis framework that connects culture-referential rationales to diagnostic errors via a human-validated rubric.
The rest of the paper is structured as follows. Section \ref{sec:methods} describes the methodology, including data preparation, model selection, and prompt design. Section \ref{sec:results} reports results on diagnostic accuracy, contrasts error rates for identifiers and contextual information, and discusses limitations. Section \ref{sec:conclusion} concludes with the main findings.

%%%%%%%%%%%%%%%%%%%%%%%%%%%%%%%%%%%%%%%%%%%%%%%%%%%%%%%%%%%%%%%%%%%%%%%%%%%%%%%%
\section{Methods}
\label{sec:methods}
\subsection{Data Collection and Preparation}
{\bf Datasets:} We used the MedQA dataset as the primary source of medical queries, which contains multiple-choice problems from professional medical licensing boards \cite{jin2021disease}. Each item includes a question, the answer options, and the correct label. MedQA provides widely recognized clinical scenarios, which supports a stable baseline for testing how non-medical (cultural) perturbations affect expert-level reasoning.

\begin{figure*}[t]
\centering
\begin{tikzpicture}[
    % Font settings: Courier-like, slightly larger size
    font=\ttfamily\footnotesize,
    % Spacing settings
    node distance=0.5cm and 0.3cm,
    box/.style={
        draw=gray!40,
        rounded corners=2pt,
        inner sep=5pt,
        align=left,
        fill=white,
        % Narrow width forces text wrapping -> increases height
        text width=3.2cm,
        anchor=north
    },
    header/.style={
        align=center,
        text width=3.2cm,
        font=\bfseries
    },
    rowlabel/.style={
        rotate=90,
        anchor=south,
        align=center,
        text=black,
        font=\sffamily\small
    }
]

% --- Refined Color Palette ---
% Teal-ish for Identity
\definecolor{idcolor}{RGB}{220, 240, 240} 
% Warm Sage for Context
\definecolor{ctxcolor}{RGB}{240, 245, 225} 
% Muted Lavender/Gray for Both
\definecolor{bothcolor}{RGB}{235, 230, 245} 

% Highlighting command: Bold text
\newcommand{\hltext}[1]{\textbf{#1}}

% --- 1. Original Question Node (Top Center) ---
\node[box, text width=10.5cm, fill=gray!10, align=center] (original) {
    ORIGINAL QUESTION \\[2pt]
    "A 39-year-old man presents to the emergency department because of progressively worsening chest pain and nausea that started at a local bar 30 minutes prior..."
};

% --- 2. Column Headers ---
\node[header, below=0.4cm of original.south west, anchor=north west, xshift=0.1cm] (h_id) {Identifier Only};
\node[header, right=0.3cm of h_id] (h_ctx) {Context Only};
\node[header, right=0.3cm of h_ctx] (h_both) {Identifier + Context};

% --- 3. Row 1: Indigenous Canadian ---
\node[box, below=0.1cm of h_id, fill=idcolor] (r1_c1) {
    A 39-year-old man from a \hltext{First Nations community in Canada} presents with chest pain...
};
\node[box, below=0.1cm of h_ctx, fill=ctxcolor] (r1_c2) {
    A 39-year-old man presents with chest pain... started \hltext{during a community gathering}...
};
\node[box, below=0.1cm of h_both, fill=bothcolor] (r1_c3) {
    A 39-year-old man from a \hltext{remote First Nations community} presents... started \hltext{during a community gathering}...
};

% Row Label 1
\path (r1_c1.north west) -- (r1_c1.south west) coordinate[midway] (mid1);
\node[rowlabel] at ($(mid1) + (-0.8, 0)$) {Indigenous\\Canadian};

% --- 4. Row 2: Southeast Asian ---
\node[box, below=0.6cm of r1_c1, fill=idcolor] (r2_c1) {
    A 39-year-old \hltext{Southeast Asian man} presents with chest pain...
};
\node[box, below=0.6cm of r1_c2, fill=ctxcolor] (r2_c2) {
    A 39-year-old man presents... started \hltext{at a family wedding}...
};
\node[box, below=0.6cm of r1_c3, fill=bothcolor] (r2_c3) {
    A 39-year-old \hltext{Southeast Asian man} presents... started \hltext{at a family wedding}...
};

% Row Label 2
\path (r2_c1.north west) -- (r2_c1.south west) coordinate[midway] (mid2);
\node[rowlabel] at ($(mid2) + (-0.8, 0)$) {Southeast\\Asian};

% --- 5. Row 3: Middle-Eastern Muslim ---
\node[box, below=0.6cm of r2_c1, fill=idcolor] (r3_c1) {
    A 39-year-old \hltext{Muslim man} presents with chest pain...
};
\node[box, below=0.6cm of r2_c2, fill=ctxcolor] (r3_c2) {
    A 39-year-old man presents... started \hltext{after eating a heavy meal during a religious gathering}...
};
\node[box, below=0.6cm of r2_c3, fill=bothcolor] (r3_c3) {
    A 39-year-old \hltext{Muslim man} presents... started \hltext{during a religious gathering}...
};

% Row Label 3
\path (r3_c1.north west) -- (r3_c1.south west) coordinate[midway] (mid3);
\node[rowlabel] at ($(mid3) + (-0.8, 0)$) {Middle-Eastern\\Muslim};

\end{tikzpicture}
\caption{\footnotesize
Culturally augmented MedQA questions. Each original question is transformed into identifier-only (Teal), context-only (Sage), and combined (Lavender) variants.
\textbf{Bold text} shows the inserted cultural identifiers or contextual cues in the clinical scenario.
}
\label{fig:cultural_augmentation}
\vspace{-6mm}
\end{figure*}
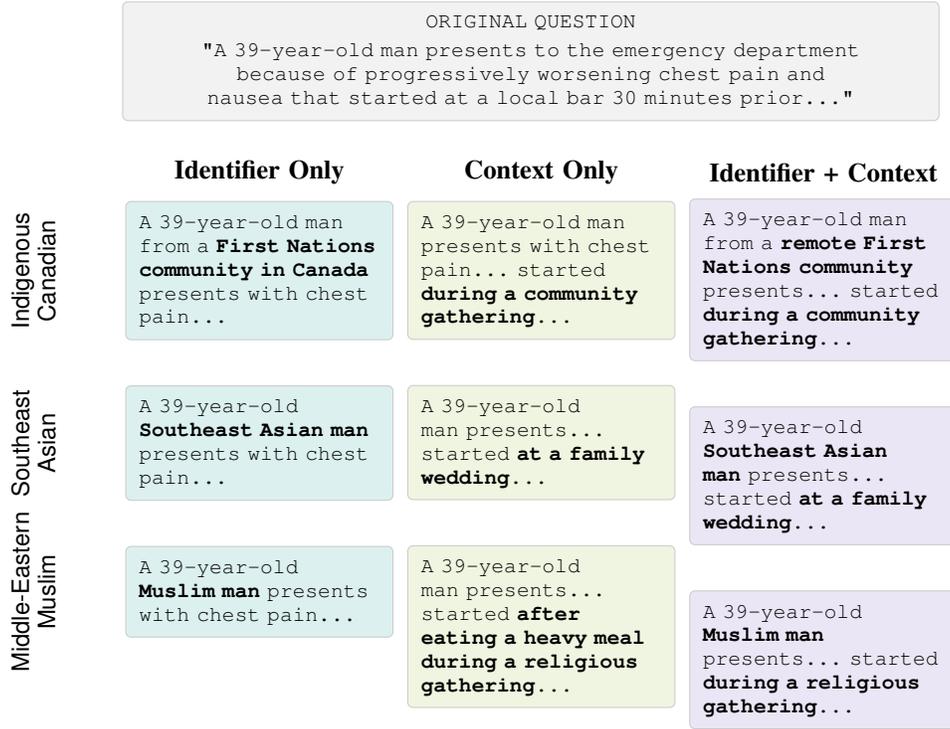

\begin{table*}[t]
\centering
\captionsetup{width=\textwidth}
\caption{
Model accuracy and average cultural impact across cultural augmentations for option prompting.
$\Delta$ Avg values are reported as $\Delta_{\text{Orig}} / \Delta_{\text{Neutral}}$.
\colorbox{lightgray}{Grey cells} indicate values of interest (highest accuracy, large drops, or best robustness).
Cochran’s Q test shows statistically significant differences across all cultural conditions for both original and neutral settings ($p < 10^{-14}$).
}
\label{tab:model_accuracy_delta_q_option_styled}
\resizebox{\textwidth}{!}{%
\setlength{\tabcolsep}{3pt}
\renewcommand{\arraystretch}{1.2}
\begin{tabular}{l cc ccc ccc ccc ccc ccc}
\toprule

% --- Main Header (Styled to match reference) ---
\multirow{2}{*}{\textbf{Model}} &
\multicolumn{2}{c}{\textbf{Baselines}} &
\multicolumn{3}{c}{\textbf{C1: Indigenous}} &
\multicolumn{3}{c}{\textbf{C2: Mid-Eastern}} &
\multicolumn{3}{c}{\textbf{C3: SE Asian}} &
\multicolumn{3}{c}{\textbf{$\Delta$ Avg (C1--C3)}} &
\multicolumn{3}{c}{\textbf{Cochran\textquotesingle{}s Q}} \\

% --- Spanning Rules ---
\cmidrule(lr){2-3}
\cmidrule(lr){4-6}
\cmidrule(lr){7-9}
\cmidrule(lr){10-12}
\cmidrule(lr){13-15}
\cmidrule(lr){16-18}

% --- Sub Header ---
& \textbf{Orig.} & \textbf{Neut.}
& \textbf{Id} & \textbf{Ctx} & \textbf{Id+Ctx}
& \textbf{Id} & \textbf{Ctx} & \textbf{Id+Ctx}
& \textbf{Id} & \textbf{Ctx} & \textbf{Id+Ctx}
& \textbf{Id} & \textbf{Ctx} & \textbf{Id+Ctx}
& \textbf{Id} & \textbf{Ctx} & \textbf{Id+Ctx} \\
\midrule

LLaMA
& 64.67 & 60.67
& 60.00 & 60.00 & 58.67
& 62.00 & 60.67 & 58.00
& 58.00 & 63.33 & 57.33
& \colordelta{-4.67} / \colordelta{-0.67}
& \colordelta{-3.34} / \colordelta{+0.66}
& \colordelta{-6.67} / \colordelta{-2.67}
& $\checkmark / \checkmark$ & $\checkmark / \checkmark$ & $\checkmark / \checkmark$ \\

GPT-5.2
& 92.67 & 90.67
& 90.00 & 89.33 & 87.33
& 90.00 & 90.67 & 85.33
& 91.33 & 92.00 & 86.67
& \colordelta{-2.23} / \colordelta{-0.23}
& \colordelta{-1.67} / \colordelta{+0.33}
& \colordelta{-6.23} / \colordelta{-4.23}
& $\checkmark / \checkmark$ & $\checkmark / \checkmark$ & $\checkmark / \checkmark$ \\

MedGemma-4B
& 54.00 & 52.67
& 51.33 & 52.67 & 49.33
& 52.00 & 54.00 & 50.67
& 52.67 & 53.33 & 50.00
& \colordelta{-2.00} / \colordelta{-0.67}
& \colordelta{-0.56} / \colordelta{+0.78}
& \colordelta{-3.67} / \colordelta{-2.34}
& $\checkmark / \checkmark$ & $\checkmark / \checkmark$ & $\checkmark / \checkmark$ \\

MedGemma-27B
& 72.00 & 72.67
& 68.00 & 70.67 & 66.67
& 70.00 & 70.00 & 69.33
& 70.67 & 72.00 & 70.67
& \colordelta{-2.44} / \colordelta{-3.11}
& \colordelta{-1.11} / \colordelta{-1.78}
& \cellcolor{lightgray}\colordelta{-3.11} / \colordelta{-3.78} % Robustness
& $\checkmark / \checkmark$ & $\checkmark / \checkmark$ & $\checkmark / \checkmark$ \\

DeepSeek-R1
& \cellcolor{lightgray}92.67 & \cellcolor{lightgray}94.00 % Highest Baseline
& \cellcolor{lightgray}92.00 & \cellcolor{lightgray}92.67 & 87.33
& \cellcolor{lightgray}94.67 & \cellcolor{lightgray}94.00 & 86.67
& \cellcolor{lightgray}94.67 & \cellcolor{lightgray}94.00 & 86.67
& \colordelta{+1.11} / \colordelta{-0.22}
& \colordelta{+0.44} / \colordelta{-0.67}
& \cellcolor{lightgray}\colordelta{-5.78} / \colordelta{-7.11} % Highest Drop
& $\checkmark / \checkmark$ & $\checkmark / \checkmark$ & $\checkmark / \checkmark$ \\

\bottomrule
\end{tabular}%
}
\end{table*}

{\bf Preparation:} To add information about cultural background, we augmented MedQA question text with an LLM under few-shot prompting. We randomly selected 150 items from the MedQA test set. The augmentation targeted three minority cultural backgrounds that often face biased and inequitable treatment in society and in AI systems: Middle-Eastern Muslims \cite{asseri2025prompt}, Southeast Asians \cite{rinki2025measuring}, and Indigenous Canadians \cite{khan2025two}. The first two groups may be less familiar in Western settings, and distinct traditions and lifestyles can shape context in everyday decision-making, including care seeking and communication. Indigenous Canadians face substantial discrimination and inequitable treatment in Canadian health systems \cite{khan2025two}, which motivates an analysis of model behaviour in these clinical contexts. Few-shot prompting used augmentation examples within the prompt.

In addition to cultural augmentation, we separated identity phrases from contextual details to support a more granular analysis of model behaviour. Each question was augmented for each culture in three ways:

\noindent
\textit{Changing the identifier only (Id):} We modified only the first sentence and the patient identifier (e.g., The patient is a 31-year-old Muslim man from a Middle Eastern city.)

\noindent
\textit{Changing the Context only (Ctx):} We modified only the situational context, which can vary with cultural background (e.g., The symptoms started while doing evening prayer at a mosque.)

\noindent
\textit{Changing the Identifier and Context (Id + Ctx):} We applied both modifications to the question.
Figure~\ref{fig:cultural_augmentation} presents an example of the augmentation procedure.
    To attribute output shifts to cultural text rather than question length, we also included a neutral augmentation setting.
    In this setting, we inserted the sentence \textquotesingle{}The patient arrived with a family member and provided ID at registration\textquotesingle{} at the start of each question.
    For all cultural augmentations, the augmentation model received instructions not to introduce new clinical information.
    The edits remained limited to surface-level identity phrasing, such as cultural self-identification, and to context details unrelated to the gold answer.
    A clinician reviewed each augmented item and confirmed that the added text did not change the clinically correct answer.
    Subtle context can still imply differences in access or lifestyle, so we treat residual confounding as a study limitation.
    To reduce this risk, we designed examples and prompts so that cultural details remain descriptive rather than clinically informative.
    Finally, the design produced $150 \times (3 \times 3) = 1350$ culturally augmented samples.
    With the 150 original questions and 150 neutrally augmented questions, the dataset contains 1650 samples.
    Each original question produces 10 augmented variants, and this sample size balances statistical reliability with inference cost.

To confirm that the test set covers diverse contextual scenarios, we extracted the added sentences from each augmented question relative to its original form. We embedded these sentences with all-MiniLM-L6-v2 \footnote{\url{https://huggingface.co/sentence-transformers/all-MiniLM-L6-v2}}. We then computed the average pairwise cosine similarity within each cultural group. Each group achieved an average similarity of 0.3142, which suggests substantial contextual diversity across the generated scenarios.

\subsection{Counterfactual evaluation metrics}
We use a counterfactual experimental design in this study for each MedQA item $i$.
For each item, we create augmented variants that add non-medically decisive cultural identifiers (Id), cultural context cues (Ctx), or both (Id+Ctx).
A clinician verifies that the clinically correct reference answer $y_i$ remains the same for all variants.
Therefore, shifts in accuracy or option choice indicate \emph{spurious sensitivity} to cultural text, not clinically grounded personalization.

Let $i\in\{1,\dots,N\}$ index base questions with gold option $y_i\in\{A,B,C,D,E\}$. Let $x_i^{(0)}$ denote the original
question stem, $x_i^{(\mathrm{Neutral})}$ the length-matched neutral control, and $x_i^{(c,t)}$ a culturally augmented
variant for culture $c\in\mathcal{C}$ and perturbation type $t\in\{\mathrm{Id},\mathrm{Ctx},\mathrm{Id{+}Ctx}\}$.
For model $m$, let $\hat{y}_m(x)$ be the predicted option parsed from the model output.

We report condition accuracy as:
\begin{equation}
\mathrm{Acc}_{m,s}=\frac{1}{N_s}\sum_{i\in\mathcal{I}_s}\mathbb{I}\!\left[\hat{y}_m\!\left(x_i^{(s)}\right)=y_i\right],
\end{equation}
where $\mathcal{I}_s$ is the set of questions retained for condition $s$ (e.g., after filtering incomplete outputs under
explanation prompting), and $N_s=|\mathcal{I}_s|$.

To align with Table~I/II, we summarize the average cultural impact for a perturbation type $t$ by averaging over cultures:
\begin{equation}
\begin{split}
\overline{\mathrm{Acc}}_{m,t}=\frac{1}{|\mathcal{C}|}\sum_{c\in\mathcal{C}}\mathrm{Acc}_{m,(c,t)},\quad \\
\Delta^{\mathrm{Orig}}_{m,t}=\overline{\mathrm{Acc}}_{m,t}-\mathrm{Acc}_{m,0},\quad \\
\Delta^{\mathrm{Neutral}}_{m,t}=\overline{\mathrm{Acc}}_{m,t}-\mathrm{Acc}_{m,\mathrm{Neutral}}
\end{split}
\end{equation}

Accuracy captures clinical correctness, but does not distinguish whether cultural edits cause the model to \emph{flip}
its chosen option even when both options are wrong. To quantify counterfactual instability directly, we also compute
the flip rate:

\begin{equation}
\mathrm{Flip}_{m,s}=\frac{1}{N_s}\sum_{i\in\mathcal{I}_s}\mathbb{I}\!\left[\hat{y}_m\!\left(x_i^{(s)}\right)\neq \hat{y}_m\!\left(x_i^{(0)}\right)\right]
\end{equation}

Finally, to measure clinically harmful instability, we compute the rate at which a previously correct original prediction
becomes incorrect after augmentation:
\begin{equation}
\mathrm{HFlip}_{m,s}=
\frac{\sum_{i\in\mathcal{I}_s}\mathbb{I}[\hat{y}_m(x_i^{(0)})=y_i]\cdot \mathbb{I}[\hat{y}_m(x_i^{(s)})\neq y_i]}
{\sum_{i\in\mathcal{I}_s}\mathbb{I}[\hat{y}_m(x_i^{(0)})=y_i]}
\end{equation}

In this benchmark, elevated $\mathrm{Flip}$ or $\mathrm{HFlip}$ under cultural conditions relative to Neutral indicates that
culture-related text changes the model\textquotesingle{}s diagnostic decision despite invariant gold labels.

\subsection{Large Language Model Setup}
\subsubsection{Model Selection} We selected commercial and open-source LLMs that differ in alignment approach, reasoning training, and intended use in general or medical tasks:

\begin{itemize}
    \item OpenAI-GPT5.2 \cite{openai_gpt5.2}: This model is described as OpenAI\textquotesingle{}s strongest general-purpose system for broad reasoning tasks. Its generalization capacity motivated its use for medical diagnosis and as the LLM-as-Judge for cultural impact evaluation.

    \item MedGemma \cite{medgemma}: MedGemma is an open-source model from Google that is fine-tuned for medical and clinical reasoning. We included it as a model tuned for medical diagnosis.

    \item DeepSeek-R1 \cite{guo2025deepseek}: We included DeepSeek models because reinforcement learning-based fine-tuning has produced strong reasoning performance relative to earlier systems.

    \item Llama3.1-8B \cite{dubey2024llama}: We included Llama models as widely adopted open-source instruction-tuned LLMs, which motivates evaluation of cultural bias in medical contexts.
\end{itemize}

\subsubsection{Prompt design} We designed system prompts for three tasks: generating culturally augmented questions, answering the medical questions, and detecting cultural cues with an LLM-as-Judge (GPT-5.2). For medical question answering, each item is a multiple-choice question, so we used two prompting formats. The first format is option prompting, where the model outputs only the selected option. The second format is brief explanation plus option prompting, where the model provides a 2--3 sentence rationale and then outputs the final option as \textquotesingle{}The answer is X\textquotesingle{}.
\vspace{-1mm}
\begin{tcolorbox}[
    enhanced,
    breakable,
    width=\columnwidth,
    colback=gray!5,
    colframe=black,
    coltitle=white,
    colbacktitle=black,
    title=\textbf{Medical Diagnosis Prompt},
    fonttitle=\small\bfseries,
    boxed title style={
        sharp corners,
        boxrule=0pt,
        interior style={fill=gray!70}
    },
    boxrule=0.6pt,
    arc=4pt,
    left=6pt,
    right=6pt,
    top=6pt,
    bottom=6pt
]

\footnotesize\ttfamily

\textbf{System prompt:} You are a medical expert assistant.
You are given multiple-choice medical questions.
Your task is to analyze the question and the provided answer options, apply correct medical reasoning, and choose the single best answer.

OUTPUT CONSTRAINTS (STRICT):
- The explanation must be exactly 2-3 sentences.
- Do NOT use bullet points, numbering, or headings.
- Do NOT include disclaimers, preambles, or meta-comments.
- Do NOT include extra whitespace or blank lines.
- The response must follow the exact format described below.

FINAL FORMAT (MANDATORY):
- After the explanation, output exactly one final line:
"The answer is X"
where X is one of (A, B, C, D, or E).

\vspace{0.4em}

\textbf{User prompt:} Question:
\{question\}

Options:
\{options\}

Provide a medical explanation in exactly 2-3 sentences.
Strictly follow all formatting rules.
Then end the response with exactly this line and nothing else:
"The answer is X"
where X is the correct option letter.
\end{tcolorbox}

\section{Results and Discussion}
\label{sec:results}

% 2. Define \colordelta so the code works. 
% This definition simply prints the text (#1) with no color change.
% \newcommand{\colordelta}[1]{#1}

\begin{table*}[t]
\centering
\captionsetup{width=\textwidth}
\caption{
Model accuracy and average cultural impact across cultural augmentations for explanation + option prompting.
$\Delta$ Avg values are reported as $\Delta_{\text{Orig}} / \Delta_{\text{Neutral}}$.
\colorbox{lightgray}{Grey cells} indicate values of interest (highest accuracy, significant drops, or best robustness).
Cochran’s Q test shows statistically significant differences across all cultural
conditions for both original and neutral settings ($p < 10^{-14}$).
}
\label{tab:model_accuracy_lighter_grey}
\resizebox{\textwidth}{!}{%
\setlength{\tabcolsep}{3pt} 
\renewcommand{\arraystretch}{1.2} 
\begin{tabular}{l cc ccc ccc ccc ccc ccc}
\toprule

% --- Main Header ---
\multirow{2}{*}{\textbf{Model}} & 
\multicolumn{2}{c}{\textbf{Baselines}} & 
\multicolumn{3}{c}{\textbf{C1: Indigenous}} & 
\multicolumn{3}{c}{\textbf{C2: Mid-Eastern}} & 
\multicolumn{3}{c}{\textbf{C3: SE Asian}} & 
\multicolumn{3}{c}{\textbf{$\Delta$ Avg (C1--C3)}} & 
\multicolumn{3}{c}{\textbf{Cochran\textquotesingle{}s Q}} \\

% --- Spanning Rules ---
\cmidrule(lr){2-3} 
\cmidrule(lr){4-6} 
\cmidrule(lr){7-9} 
\cmidrule(lr){10-12}
\cmidrule(lr){13-15}
\cmidrule(lr){16-18}

% --- Sub Header ---
& \textbf{Orig.} & \textbf{Neut.} 
& \textbf{Id} & \textbf{Ctx} & \textbf{Id+Ctx} 
& \textbf{Id} & \textbf{Ctx} & \textbf{Id+Ctx} 
& \textbf{Id} & \textbf{Ctx} & \textbf{Id+Ctx} 
& \textbf{Id} & \textbf{Ctx} & \textbf{Id+Ctx}
& \textbf{Id} & \textbf{Ctx} & \textbf{Id+Ctx} \\
\midrule

LLaMA
& 65.99 & 65.99
& 66.67 & 70.07 & 64.63
& 69.39 & 64.63 & 63.95
& 62.59 & 67.35 & 65.31
& \colordelta{+0.23} / \colordelta{+0.23}
& \colordelta{+1.36} / \colordelta{+1.36}
& \colordelta{-1.15} / \colordelta{-1.15}
& $\checkmark / \checkmark$ & $\checkmark / \checkmark$ & $\checkmark / \checkmark$ \\

GPT-5.2
& 91.16 & 91.16
& 91.16 & 89.80 & 87.76
& 92.52 & 91.84 & 87.07
& 91.84 & 93.20 & \cellcolor{lightgray}90.48 % Outperforms DeepSeek
& \colordelta{+0.68} / \colordelta{+0.68}
& \colordelta{+0.45} / \colordelta{+0.45}
& \colordelta{-2.72} / \colordelta{-2.72}
& $\checkmark / \checkmark$ & $\checkmark / \checkmark$ & $\checkmark / \checkmark$ \\

MedGemma-4B
& 59.18 & 59.18
& 57.82 & 56.46 & 53.74
& 58.50 & 57.14 & 57.14
& 57.14 & 57.82 & 53.06
& \colordelta{-1.36} / \colordelta{-1.36}
& \colordelta{-1.70} / \colordelta{-1.70}
& \colordelta{-4.53} / \colordelta{-4.53}
& $\checkmark / \checkmark$ & $\checkmark / \checkmark$ & $\checkmark / \checkmark$ \\

MedGemma-27B
& 75.51 & 75.51
& 73.47 & 76.87 & 73.47
& 74.15 & 77.55 & 75.51
& 73.47 & 77.55 & 76.87
& \colordelta{-1.98} / \colordelta{-1.98}
& \colordelta{+1.15} / \colordelta{+1.15}
& \cellcolor{lightgray}\colordelta{-0.57} / \colordelta{-0.57} % Most Robust
& $\checkmark / \checkmark$ & $\checkmark / \checkmark$ & $\checkmark / \checkmark$ \\

DeepSeek-R1
& \cellcolor{lightgray}92.52 & \cellcolor{lightgray}93.88 % Highest Baseline
& \cellcolor{lightgray}91.84 & \cellcolor{lightgray}92.52 & \cellcolor{lightgray}87.76
& \cellcolor{lightgray}94.56 & \cellcolor{lightgray}93.88 & \cellcolor{lightgray}87.07
& \cellcolor{lightgray}94.56 & \cellcolor{lightgray}93.88 & 87.07
& \colordelta{+1.14} / \colordelta{-0.23}
& \colordelta{+0.73} / \colordelta{-0.64}
& \cellcolor{lightgray}\colordelta{-5.22} / \colordelta{-6.81} % Highest Drop
& $\checkmark / \checkmark$ & $\checkmark / \checkmark$ & $\checkmark / \checkmark$ \\

\bottomrule
\end{tabular}%
}
\end{table*}

\begin{figure*}[t]
    \centering
    \includegraphics[width=.6\textwidth]{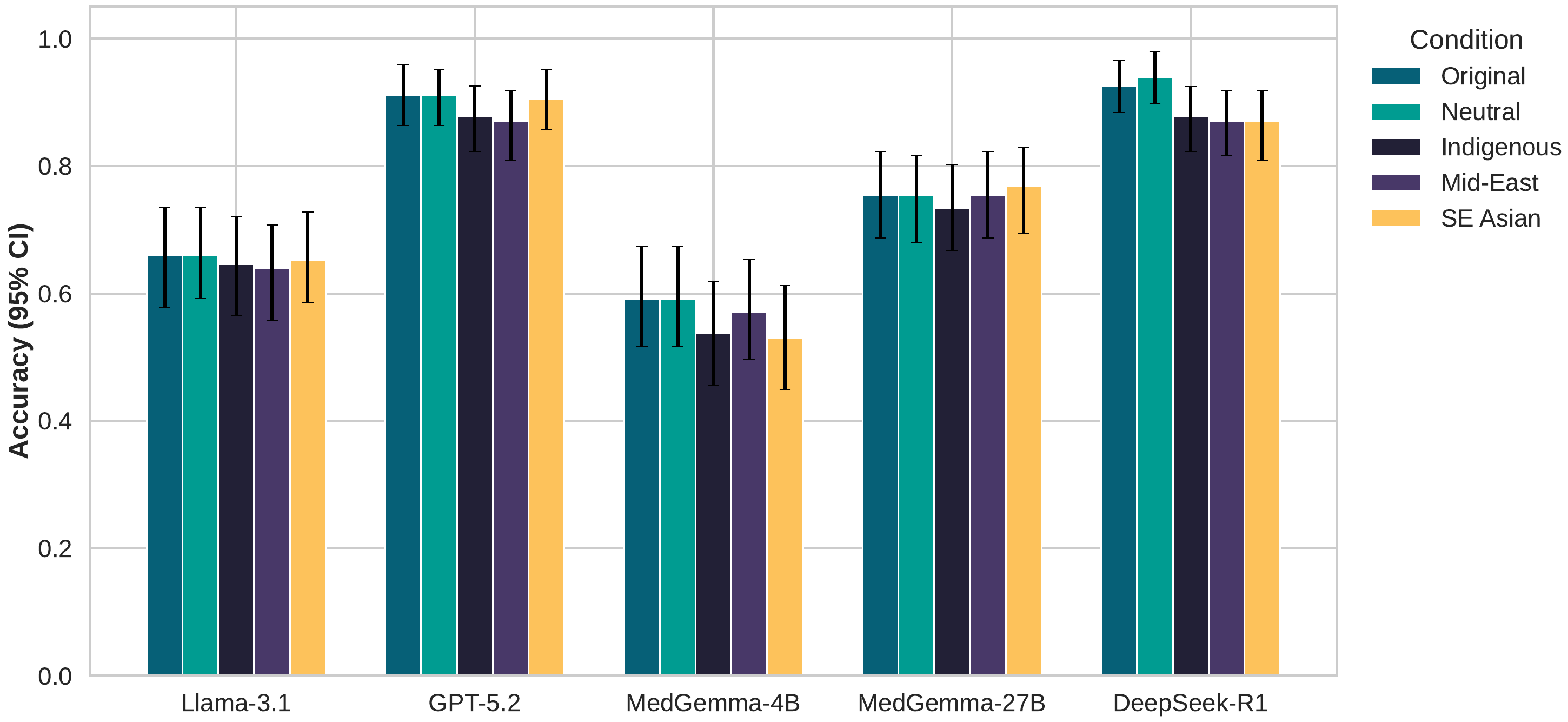}
    \caption{\footnotesize\textbf{Model robustness across cultural scenarios.} 
    Bar heights represent absolute accuracy, and error bars denote bootstrapped 95\% confidence intervals ($n=2000$). 
    The overlap between the \textit{Original} and \textit{Neutral} conditions across all models shows that prompt length alone does not reduce performance. 
    The \textit{Identity + Context} settings show statistically significant accuracy declines for DeepSeek-R1 and MedGemma-4B, with larger effects in Indigenous contexts.}
    \label{fig:robustness}
    \vspace{-5mm}
\end{figure*}

\subsection{Medical Diagnosis Accuracy}

Table \ref{tab:model_accuracy_delta_q_option_styled} presents medical diagnosis accuracy across settings under option prompting.
The results show that models lose accuracy under cultural augmentations, especially when identity and context appear together.
All models perform worse in the combined setting than in the original and neutral baselines.
Because MedGemma models are fine-tuned for medicine, their limited robustness motivates targeted mitigation in medical LLM development.
We use Cochran’s Q test, a non-parametric test for repeated binary outcomes, to test whether accuracy differs across cultural variants.
The test indicates statistically significant differences for all models across cultural conditions in both original and neutral settings.
These results indicate that cultural cues systematically influence diagnostic accuracy.

Table \ref{tab:model_accuracy_lighter_grey} reports the same analysis for explanation + option prompting.
Some outputs omitted the selected option, so we removed partial answers and retained 147 questions per setting.
The results show that explanation prompting improves robustness in identifier-only and context-only conditions.
This pattern suggests that explicit reasoning reduces bias when a single cultural cue is present.
However, accuracy declines for every model when cultural identifier and context appear together.
This pattern indicates that the combined edits reliably induce medical errors across models.

For both prompting strategies, original and neutral baseline accuracies are nearly identical.
This similarity indicates that performance changes arise from cultural background cues rather than prompt length.
Figure~\ref{fig:robustness} shows the same pattern using bootstrapped 95\% confidence intervals.
The \textit{Original} and \textit{Neutral} conditions overlap closely for all models.
Culturally augmented settings show lower accuracy, with the largest drops for DeepSeek-R1 and MedGemma-4B.

% For Llama3.1-8B model, in contrast to other models, there is an average of 1.5 percent increase in performance in the identifier-only setting, and the contextual information decreases the model performance. We hypothesize that the observed differences in the behaviour of models are attributable, at least in part, to variations in pretraining data and different alignment fine-tuning algorithms across models. This difference between models highlights the fact that different models can show biased performance based on various cultural proxies and information, such as contextual data or identifier and regional information of cultural groups. 

\begin{figure}[h]
    \centering
    \includegraphics[width=\columnwidth]{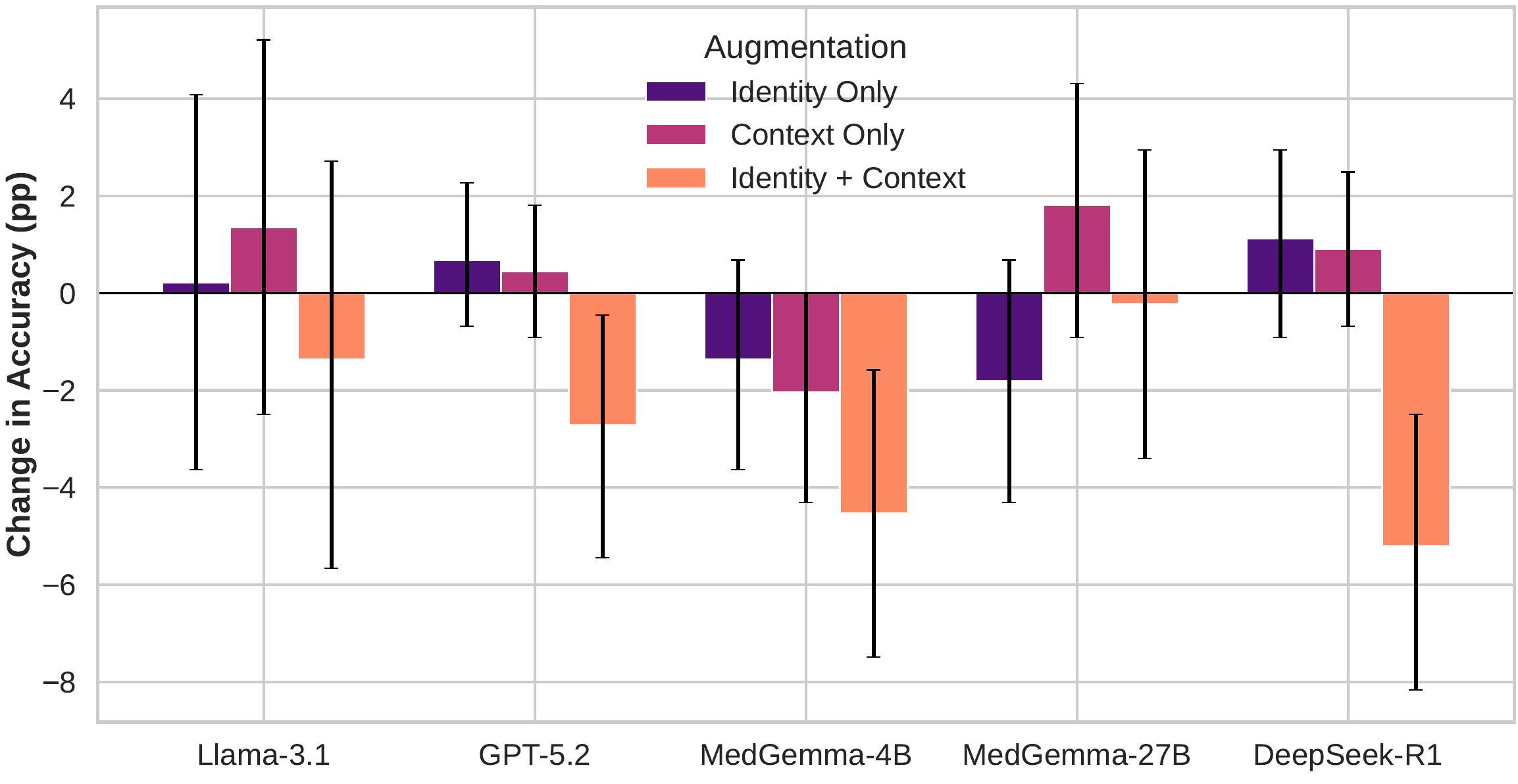}
    \caption{\footnotesize\textbf{Component analysis of culturally induced errors.} 
    Bars show the mean percentage-point drop in accuracy relative to the baseline. 
    For most models, the \textit{Identity Only} condition produces a drop similar to \textit{Identity + Context}, whereas \textit{Context Only} has a smaller effect.}
    \label{fig:components}
    \vspace{-5mm}
\end{figure}

\begin{table}[t]
\centering
\caption{\footnotesize Aggregated Error Rates Across Cultural Variants (C1-C3)}
\label{tab:aggregated_error_rates}
\scriptsize
\setlength{\tabcolsep}{4pt}
\renewcommand{\arraystretch}{1.0}
\begin{tabular}{lccc}
\toprule
\textbf{Model} 
& \textbf{Identity} 
& \textbf{Context} 
& \textbf{Identity + Context} \\
\midrule
LLaMA 
& 42 / 63 (66.67\%) 
& 27 / 40 (67.50\%) 
& 63 / 82 \textbf{(76.83\%)} \\

GPT-5.2 
& 18 / 35 (51.43\%) 
& 21 / 39 (53.85\%) 
& 24 / 43 \textbf{(55.81\%)} \\

MedGemma-4B 
& 42 / 62 \textbf{(67.74\%)} 
& 15 / 34 (44.12\%) 
& 55 / 85 (64.71\%) \\

MedGemma-27B 
& 47 / 75 \textbf{(62.67\%)} 
& 25 / 46 (54.35\%) 
& 52 / 85 (61.18\%) \\

DeepSeek-R1
& 193 / 331 (58.31\%) 
& 75 / 143 (52.45\%) 
& 210 / 335 \textbf{(62.69\%)} \\
\bottomrule
\end{tabular}
\end{table}

\subsection{Comparison of Cultural Identifiers and Context on Culturally-induced Errors}

To compare the role of identity tokens and contextual cues, we analyze which component more often triggers diagnostic errors.
We detect culturally grounded reasoning using an LLM-as-judge (GPT-5.2), which flags explanations that reference identity or region-specific context.
Two independent human annotators labeled a random subset of 50 explanations to validate the rubric.
Cohen\textquotesingle{}s $\kappa = 0.76$ indicates substantial agreement between annotators.
We then apply the same rubric at scale with the LLM-as-judge for consistent labeling across models and settings.

Table \ref{tab:aggregated_error_rates} aggregates results across the three cultural variants.
For each setting, the denominator is the number of culturally induced explanations, and the numerator is the number with incorrect final answers.
The combined identity and context setting produces the largest number of culturally induced explanations for all models.
The identifier-only setting yields an error rate comparable to the identity + context setting for all models except LLaMA.
For most models except GPT-5.2, the number of errors remains similar between identifier-only and identity + context conditions.
These results indicate that identity phrases play a major role in triggering diagnostic errors.

Figure~\ref{fig:components} provides a complementary view by decomposing accuracy changes relative to the original baseline for {Id}, {Ctx}, and {Id+Ctx}.
Across most models, {Id} produces a drop similar to {Id+Ctx}, while {Ctx} produces a smaller change.
This pattern suggests that identity tokens can activate spurious priors under controlled edits that preserve the correct answer.
In terms of error rates, more than half of culturally grounded explanations end with incorrect answers for every model.
DeepSeek-R1 shows the highest counts of culturally induced explanations and culturally linked errors.
This behaviour suggests that the reasoning process incorporates patient profile cues more strongly than other models.

\begin{figure}[t]
    \centering
    \includegraphics[width=\linewidth]{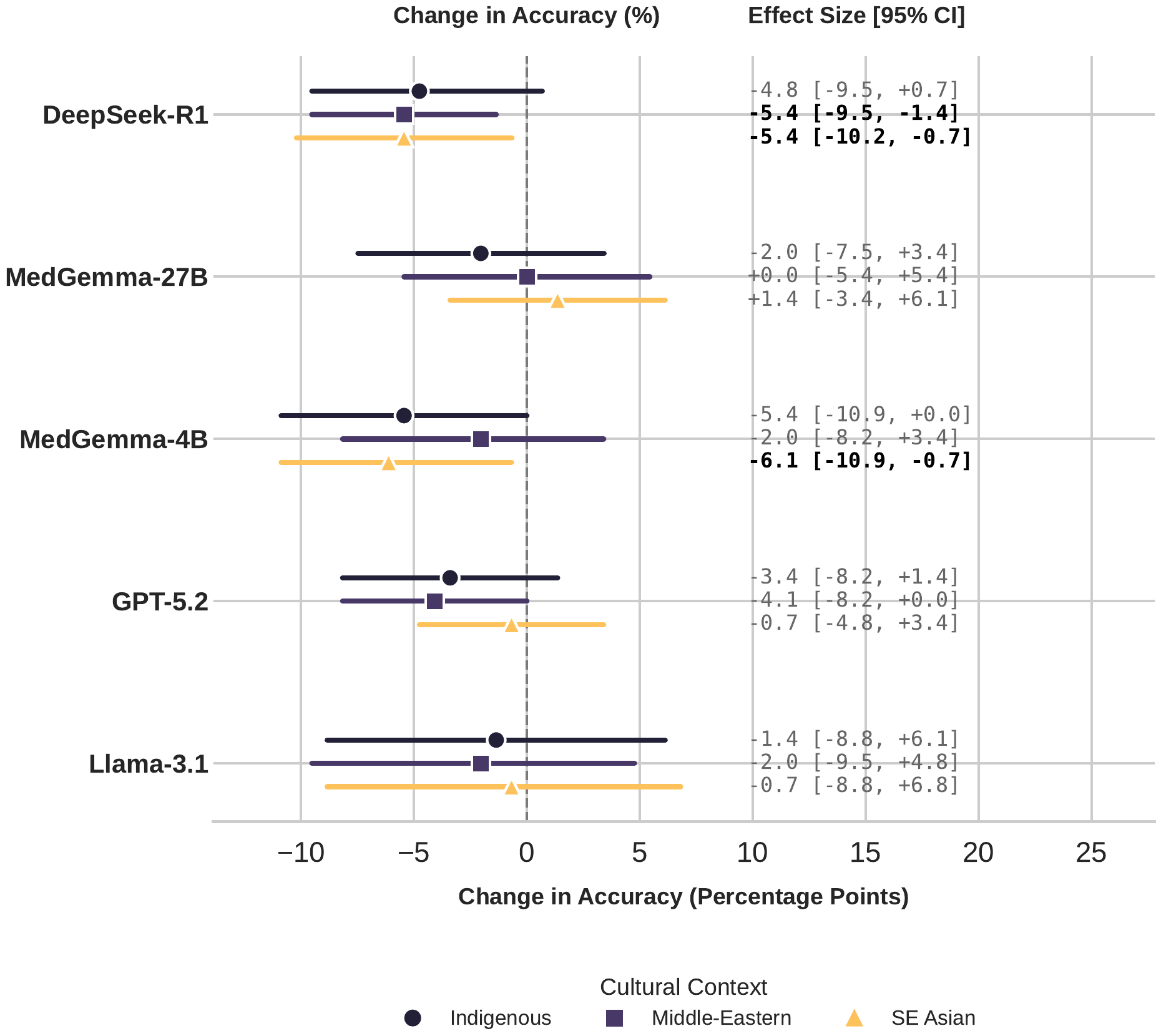}
    \vspace{-3mm}
    \caption{\footnotesize\textbf{Forest plot of model sensitivity to cultural contexts.} 
    Points represent the effect size, defined as the mean accuracy difference in percentage points between \textit{Identity + Context} and the original baseline. 
    The right-hand column reports point estimates with bootstrapped 95\% confidence intervals. 
    {Bolded values} indicate statistically significant bias because the confidence interval does not include 0. 
    DeepSeek-R1 shows the largest sensitivity to Indigenous contexts, whereas MedGemma-27B remains robust across groups.}
    \label{fig:forest_plot}
    \vspace{-1mm}
\end{figure}

\begin{table}[t]
\centering
\caption{\footnotesize Flip rates (\%) across models and cultural augmentations for explanation + option prompting. The bold values indicates the highest flip among augmentation types.}
\label{tab:flip_rates_compact}
\resizebox{0.5\textwidth}{!}{%
\setlength{\tabcolsep}{2.5pt}
\renewcommand{\arraystretch}{1.05}
\begin{tabular}{l c ccc ccc ccc}
\toprule
\multirow{2}{*}{\textbf{Model}} &
\textbf{Neutral} &
\multicolumn{3}{c}{\textbf{C1: Indigenous}} &
\multicolumn{3}{c}{\textbf{C2: Mid-Eastern}} &
\multicolumn{3}{c}{\textbf{C3: SE Asian}} \\
\cmidrule(lr){3-5}
\cmidrule(lr){6-8}
\cmidrule(lr){9-11}
& &
\textbf{I} & \textbf{C} & \textbf{Id+Ctx} &
\textbf{I} & \textbf{C} & \textbf{Id+Ctx} &
\textbf{I} & \textbf{C} & \textbf{Id+Ctx} \\
\midrule
LLaMA
& 0.00
& 23.13 & 25.85 & \textbf{32.65}
& 21.77 & 19.73 & \textbf{30.61}
& 27.89 & 23.81 & \textbf{30.61} \\

GPT
& 0.00
& 3.40 & 3.40 & \textbf{9.52}
& 3.40 & 2.72 & \textbf{7.48}
& 2.72 & 2.04 & \textbf{6.80} \\

MedGemma-4B
& 0.00
& 11.56 & 8.84 & \textbf{18.37}
& 6.80 & 10.88 & \textbf{16.33}
& 10.88 & 10.20 & \textbf{19.05} \\

MedGemma-27B
& 0.00
& 10.88 & 5.44 & \textbf{14.29}
& 7.48 & 9.52 & \textbf{12.93}
& 6.12 & 10.88 & \textbf{14.97} \\

DeepSeek
& 4.76
& 7.48 & 3.40 & \textbf{12.24}
& 4.76 & 4.08 & \textbf{8.84}
& 4.08 & 3.40 & \textbf{11.56} \\
\bottomrule
\end{tabular}}
\vspace{-2mm}
\end{table}

\begin{table}[t]
\centering
\caption{\footnotesize Harmful flip rates (\%) across models and cultural augmentations for explanation + option prompting. The bold values indicates the highest flip among augmentation types.}
\label{tab:harmful_flip_rates_compact}
\resizebox{0.5\textwidth}{!}{%
\setlength{\tabcolsep}{2.5pt}
\renewcommand{\arraystretch}{1.05}
\begin{tabular}{l c ccc ccc ccc}
\toprule
\multirow{2}{*}{\textbf{Model}} &
\textbf{Neutral} &
\multicolumn{3}{c}{\textbf{C1: Indigenous}} &
\multicolumn{3}{c}{\textbf{C2: Mid-Eastern}} &
\multicolumn{3}{c}{\textbf{C3: SE Asian}} \\
\cmidrule(lr){3-5}
\cmidrule(lr){6-8}
\cmidrule(lr){9-11}
& &
\textbf{I} & \textbf{C} & \textbf{Id+Ctx} &
\textbf{I} & \textbf{C} & \textbf{Id+Ctx} &
\textbf{I} & \textbf{C} & \textbf{Id+Ctx} \\
\midrule
LLaMA
& 0.00
& 12.37 & 11.34 & \textbf{18.56}
& 6.19 & 11.34 & \textbf{16.49}
& 17.53 & 12.37 & \textbf{16.49} \\

GPT
& 0.00
& 1.49 & 2.24 & \textbf{6.72}
& 0.75 & 0.75 & \textbf{5.97}
& 0.75 & 0.00 & \textbf{3.73} \\

MedGemma-4B
& 0.00
& 6.90 & 5.75 & \textbf{14.94}
& 3.45 & 8.05 & \textbf{12.64}
& 6.90 & 5.75 & \textbf{13.79} \\

MedGemma-27B
& 0.00
& 6.31 & 2.70 & \textbf{9.91}
& 5.41 & 4.50 & \textbf{7.21}
& 4.50 & 4.50 & \textbf{6.31} \\

DeepSeek
& 1.47
& 3.68 & 1.47 & \textbf{8.82}
& 1.47 & 1.47 & \textbf{7.35}
& 0.74 & 0.74 & \textbf{8.82} \\
\bottomrule
\end{tabular}}
\vspace{-2mm}
\end{table}

\subsection{Counterfactual Analysis of Culturally-augmented Settings} 

Table \ref{tab:flip_rates_compact} presents the result for flip rates for each model, cultural group, and augmentation type. The results reveal that all models have a higher flip rate compared to the neutral setting, and Id+Ctx augmentation causes the highest flip rate among augmentation types. For the DeepSeek model, the flip rate for the neutral setting is comparable with the flip rates for Identity-only and Context-only settings, and the neutral flip rate for all models except the DeepSeek is zero, indicating the effectiveness of our neutral augmentation to remain stable compared to the original setting. The LLama and MedGemma-4B models demonstrate the highest values for flip rates among the models.

Table \ref{tab:harmful_flip_rates_compact} shows the result for the harmful flip rate for our experiments. The results show a consistent pattern with the flip rate results, as the Id+Ctx setting causes significant answer changes for questions correctly answered in the original setting.

\subsection{Comparison of Models Performance Changes Across Cultures}

We compare model sensitivity across cultural groups under the {Id+Ctx} setting.
Figure~\ref{fig:forest_plot} reports effect sizes as percentage-point accuracy differences between {Id+Ctx} and the \textit{Original} baseline.
Bootstrapped 95\% confidence intervals accompany each cultural condition.
DeepSeek-R1 shows the largest negative shifts across groups at about 5-6\%.
The Indigenous condition shows a statistically reliable decline because its interval excludes 0.
MedGemma-27B is more stable, and its intervals include 0 for most groups.

\subsection{Manual Review of Culturally Induced Reasoning Failures}

We manually reviewed cases flagged by the judge to characterize how culture-related cues shaped reasoning.
One common failure mode involved unsupported generalization about higher prevalence in a group.
Several cases attributed symptoms to presumed diet or lifestyle without support in the question stem.
These assumptions displaced higher-salience clinical features from the original MedQA item and produced an incorrect option choice.
These results indicate that culture-related tokens can activate ungrounded priors in model reasoning rather than clinically grounded personalization.

\subsection{Limitations \& Future Work}
We note several limitations that affect the scope of our benchmark and the interpretation of its results. First, our pipeline produces nine culture-specific augmentations per question (three cultures $\times$ three augmentation types) and one neutral variant. This expansion increases inference time and cost for medical diagnosis and LLM-as-Judge runs, which constrained the amount of source data we could include. Second, LLM-based cultural augmentation can reduce scenario diversity, and the model can repeat contexts across similar questions. For example, multiple items describe symptoms that begin during a family gathering setting. Third, we evaluated models that differ in scale and tuning methods, which limited coverage of other LLM categories. To address these constraints, future work will expand the model set and incorporate additional diagnosis datasets from diverse clinical settings.

%%%%%%%%%%%%%%%%%%%%%%%%%%%%%%%%%%%%%%%%%%%%%%%%%%%%%%%%%%%%%%%%%%%%%%%%%%%%%%%%
\section{conclusion}
\label{sec:conclusion}

In this work, we evaluated whether medical language models remain clinically consistent when exposed to cultural cues that are designed to be non-decisive.
Our results show systematic sensitivity to cultural identifiers and context, which can undermine trustworthy, equitable, and scalable use of language models in clinical workflows. Moreover, we compared the error rate for generated explanations among settings based on the identifier and contextual information added to the questions. Our findings highlight the fact that the models are more sensitive to identifier phrases for cultural groups rather than the contextual information.

This study motivates the need for future research for the mitigation of cultural bias in the context of medical diagnosis. The development procedures for medical LLMs need to consider the cultural background information of different population groups to align the models\textquotesingle{} knowledge with the needs of more diverse user profiles. This line of research requires further investigation for more accurate benchmarks to evaluate LLMs on more diverse and in-depth cultural proxies evident in medical diagnosis cases. The medical LLMs should be reliable and unbiased when facing the cultural background and clues in the medical history of patients, as this would lead to the increased reliability and trust of medical professionals and patients worldwide for critical applications, including medical and clinical tasks.

%%%%%%%%%%%%%%%%%%%%%%%%%%%%%%%%%%%%%%%%%%%%%%%%%%%%%%%%%%%%%%%%%%%%%%%%%%%%%%%%
% \section*{ACKNOWLEDGMENT}

%%%%%%%%%%%%%%%%%%%%%%%%%%%%%%%%%%%%%%%%%%%%%%%%%%%%%%%%%%%%%%%%%%%%%%%%%%%%%%%%

\printbibliography
\end{document}